\documentclass[10pt,twocolumn]{article} 
\usepackage[left=1.5cm,right=1.5cm, top=2.00cm, bottom=2.00cm]{geometry}
\usepackage{cite}
\usepackage{authblk}
\usepackage{hyperref}

\usepackage{graphicx}
\usepackage{amsmath,amssymb,amsfonts}
\usepackage{scalerel}
\usepackage{xcolor}

\usepackage{bibspacing}
\usepackage{etoolbox}
\AtBeginEnvironment{tabularx}{\tiny} 

\usepackage[compact]{titlesec}

\renewcommand{\vec}[1]{\mathbf{#1}}
\usepackage{tabularx,booktabs,multicol,multirow}

\newcommand{\R}{\ensuremath{\mathbb{R}}}
\newcommand{\rv}{\ensuremath{\rVert_{2M}}}
\date{}
\providecommand{\keywords}[1]
{
  \small	
  \textbf{\textit{Keywords---}} #1
}

\title{\textbf{Visualisation and knowledge discovery from interpretable models}\\
\normalsize{ Preprint of IJCNN 2020 as prepared by the authors\thanks{\textcopyright   20XX IEEE.  Personal use of this material is permitted.  Permission from IEEE must be obtained for all other uses, in any current or future media, including reprinting/republishing this material for advertising or promotional purposes, creating new collective works, for resale or redistribution to servers or lists, or reuse of any copyrighted component of this work in other works}}}
\author[1]{Sreejita Ghosh}
\author[2]{Peter Tino}
\author[1]{Kerstin Bunte}
\affil[1]{\textit{Bernoulli Institute, University of Groningen, The Netherlands}}
\affil[2]{\textit{School of Computer Science, University of Birmingham, The United Kingdom}}
\begin{document}


\maketitle
\begin{abstract}
Increasing number of sectors which affect human lives, are using Machine Learning (ML) tools. Hence the need for understanding their working mechanism and evaluating their fairness in decision-making, are becoming paramount,  ushering in the era of Explainable AI (XAI).
In this contribution we introduced a few intrinsically interpretable models which are also capable of dealing with missing values, in addition to extracting knowledge from the dataset and about the problem. These models are also capable of visualisation of the classifier and decision boundaries: they are the angle based variants of Learning Vector Quantization. We have demonstrated the algorithms on a synthetic dataset and a real-world one (heart disease dataset from the UCI repository). The newly developed classifiers helped in investigating the complexities of the UCI dataset as a multiclass problem. The performance of the developed classifiers were comparable to those reported in literature for this dataset, with additional value of interpretability, when the dataset was treated as a binary class problem. 

\end{abstract}


\section{Introduction}
In this era of increasing number of machine learning (ML) algorithms being deployed in various sectors, including finance, healthcare, criminology, justice, politics, 
manufacturing, and logistics, more and more human lives are impacted by them. 
Consequently there is a rising need of transparency and interpretability of the models \cite{doshi2017towards, bibal2016interpretability, carvalho2019machine} to achieve comprehensible decisions.
ML algorithms with greater predictive powers are often more complex and behave like a 
\textit{black box}, i.e. the working logic of these models is concealed from the human experts, thus obviating any way of verifying the reasoning and thus, the fairness of system \cite{carvalho2019machine}.
However role of ML in high-stake prediction applications concerning human lives demand that its decisions be explainable by humans \cite{carvalho2019machine}. 

However, there have been debates about the meaning of the term interpretability, and how to compare interpretability of different classifiers, especially when 
comparing models of distinct types. 
To tackle this problem Backhaus and Seiffert proposed 3 criteria\cite{backhaus2014classification,bibal2016interpretability}: 
(1) the model's ability to perform feature selection from the input pattern, 
(2) the model's ability to provide typical data points representing a class, and 
(3) model parameters having information about the decision boundary directly encoded.
Different strategies have been proposed: including model-agnostic pre- or post-processing methods such as univariate feature selection \cite{carvalho2019machine} and post hoc 
visualisation of decision boundaries \cite{Bunte2011_LiRaMLVQ,schulz2015using}.
This contribution focuses on intrinsically interpretable techniques and hence model-specific examples. 
Using these criteria Support Vector Machines (SVM) models \cite{bibal2016interpretability} are graded 1 out of 3 because they satisfy only criteria (3), to contain information about the decision boundary. 
In Decision trees (DTs) \cite{benard2019sirus} rules are interpretable. 
A typically higher performance classifier, Random Forest (RF), is built by bagging several DTs on random subsets of the data. However ensembling compromises on  interpretability. 
Naive Bayes (NB) assumes independence of features which leads to interpretability of individual features and their contribution for decision making. 
However it lacks the ability to account for feature interactions for the target outcome \cite{carvalho2019machine}. 
In this paper we aim to develop a \textit{competitive} 
classifier in terms of performance, which is also easily interpretable, and can be visualised, satisfying criteria 1-3 \cite{backhaus2014classification}.

Nearest Prototype Classification (NPC) is an intuitive learning scheme where a novel sample gets assigned the class label of its closest prototype. 
Thus techniques implementing it, such as Generalized LVQ (GLVQ) \cite{sato} for example. often allow interpretation of the prototypes as representative of class information allowing transparency with respect to (2). 
The Generalized Relevance LVQ (GRLVQ) \cite{Hammer20021059} extension to it additionally provide feature relevance determination by introduction of an adaptive parameterized dissimilarity. This weighs the importance of features for the classification and makes this extension fulfill criteria (1) as well. 
Further adaptations allow for multi-variate and class-wise feature analysis \cite{Schneider2007,schneider2009adaptive} and visualisation of decision boundaries \cite{Bunte2011_LiRaMLVQ}. 
However certain datasets, such as medical data, 
often contain missing values, heterogeneous measurements, and frequently exhibit imbalanced classes which often hinder the straightforward application of ML algorithms. 

We addressed the aforementioned challenges by introducing an angular adaptive dissimilarity measure and an oversampling strategy in \cite{ghosh2017comparison}. 
In this contribution we present and demonstrate extensions to \cite{ghosh2017comparison} which allow for knowledge discovery from non-linearly separable datasets exhibiting the mentioned hindrances. The proposed interpretable classifiers are demonstrated on a synthetic and a publicly available dataset. 
These classifiers are capable of class-wise and multi-variate feature analysis and visualisation of non-linear decision boundaries (see section \ref{sec:methods}), thus satisfying at least 2 of the 3 criteria of \cite{backhaus2014classification}. 
Detailed explanation of GLVQ and its extensions relevant to this paper can be found in section \ref{sec:methods}.


\section{Methods}
\label{sec:methods}
In this section we present the interpretable LVQ algorithm capable of dealing with missingness and proposed extensions for non-linear decision boundaries and visualisation. 
%
%
We assume training is based on $S$ data samples $\{\vec{x}_i\in\R^{D}\}_{i=1}^S$ accompanied by a label $c(\vec{x}_i)$ 
belonging to one of $C$ classes and a set of adaptive prototypes 
$\vec{w}\in\R^D$ with labels $c(\vec{w})$. 
A new data sample receives a label following a prototype-based nearest neighbor classification scheme: 
by assigning the label of the closest prototype with $c(\vec{w}_J)=\arg\min_J d_i^J$ using a dissimilarity measure $d_i^J=d(\vec{x}_i,\vec{w}_J)$. 
The paper by \cite{sato} introduced Generalized LVQ (GLVQ), in which the prototype positions were optimsed using the following cost function: 
\begin{align}
\label{eq:GLVQ}
E = \sum_{i=1}^S\Phi\left(\frac{d_i^J-d_i^K}{d_i^J+d_i^K}\right) \enspace, 
\end{align}
with $d_i^J$ being the Euclidean distance of each training sample to the closest prototype of the same class $c(\vec{x}_i)=c(\vec{w}_J)$ and $d_i^K$ the closest prototype with another class label. 
$\Phi$ is a monotonic function and we set it to the identity $\Phi(a) = a$ throughout this contribution. 
Learning takes place by adapting the prototypes $\vec{w}$, e.g. by stochastic gradient descent 
updating the closest correct and wrong prototypes $\vec{w}^L$, $L\in\{J,K\}$ using the derivatives $\nabla\vec{w}^L=\frac{\partial E}{\partial\vec{w}^L}$:
\begin{align} 
\notag
\frac{\partial E}{\partial \vec{w}^J} &= \sum_{i=1}^S\gamma_i^J\frac{\partial d_i^J}{\partial \vec{w}^J} \text{ and }
&\frac{\partial E}{\partial \vec{w}^K} =& \sum_{i=1}^S\gamma_i^K\frac{\partial d_i^K}{\partial \vec{w}^K} \text{ with}\\
\gamma_i^J &=\frac{2d^K_i}{(d_i^J+d_i^K)^2} \text{ and } 
&\gamma_i^K =&\frac{-2d^J_i}{(d_i^J+d_i^K)^2}
\label{eqn:CommDerW}
\end{align}
After training the prototypes can often be considered typical representatives of their class and their characteristics can be investigated for interpretation.  

Since the Euclidean distance is sensitive to missing values the authors introduced an angle-based variant ALVQ allowing
learning in variable dimensional spaces \cite{Bunte_NC_2_2016,ghosh2017comparison}: 
\begin{align}
d_i^L &= g_\beta(b) 
= \frac{e^{(-\beta (b-1))}-1}{e^{(2\beta)}-1} 
\text{ with }b=\frac{\vec{x}_i\cdot\vec{w}^L}{\lVert\vec{x}_i\rVert\lVert\vec{w}^L\rVert}
\enspace .
\end{align}
The exponential function $g_\beta(b)$ transforms the angle $b=\cos\theta\in[-1,1]$ into dissimilarities in [0,1] with 
the hyper-parameter $\beta$ influencing the slope, e.g. $\beta\rightarrow 0$ leading to a near linear relationship. 
In presence of missing data the angle $b$ and derivatives are computed with the available dimensions only. 
Optimization takes place deriving the cost function $E$ Eq.\ (\ref{eq:GLVQ}-\ref{eqn:CommDerW}) with changed dissimilarity $d_i^L$ adding:
\begin{align} 
\frac{\partial d_i^L}{\partial \vec{w}^L} &= \frac{\partial g_\beta(b)}{\partial b} \cdot \frac{\partial b}{\partial \vec{w}^L} \text{ and}\\ 
\frac{\partial g_\beta(b)}{\partial b} & = \frac{-\beta\exp(-\beta b+\beta)}{\exp(2\beta)-1} \enspace .
\end{align}
The update rules of GLVQ contains forces attracting the closest correct prototype for each data sample and repulsion of the closest one with a different class label. 
For example in an imbalanced 2 class problem the Euclidean variant might push the minority class prototype far away from the data all together, since it is being repelled more often by the majority class 
than attracted by the minority class. 
ALVQ classifies on the hypersphere, so a prototype cannot be infinitely repelled without returning on the other side, leading to more stable behaviour facing imbalance. 
Finally, the dissimilarity measure $d_i^L$ can be parameterized leading to several powerful extensions with varying potential for further interpretation. 
We group the novel angle extensions into three categories, namely global, local and 2 matrix, as explained in the following subsections.

\subsection{Global relevance matrix}\label{angleGMLVQ} 

First extensions to GLVQ introduced parameterized dissimilarity measures based on the quadratic form:
\begin{align} 
\label{eq:d_Lambda}
d^L_i = (\vec{x}_i-\vec{w}^L)^\top\Lambda(\vec{x}_i-\vec{w}^L) \enspace,
\end{align}
with the semi-definite matrix $\Lambda\in\R^{D\times D}$ containing additional parameters for optimization. 
A variant called Relevance GLVQ (GRLVQ) \cite{Hammer20021059} assumes $\Lambda$ to be a diagonal matrix with $\sum_{i=1}^D\Lambda_{ii}^2=1$. 
The diagonal elements $r_i=\Lambda_{ii}^2$ allow learning of discriminant feature directions, which automatically reduces the influence of less relevant measurement dimensions.
However GRLVQ is univariate and does not take into account features which are relevant only in combination with another. 
Generalized Matrix LVQ (GMLVQ) \cite{Schneider2007,schneider2009adaptive,schneider2009distance} tackles this issue by allowing a full matrix $\Lambda$, ensuring semi-definiteness by the decomposition 
$\Lambda=\Omega^\top \Omega$ and optimizing $E$ with respect to $\Omega\in\R^{D\times D}$. 
Since $d^L_i$ can be rewritten as squared Euclidean distance in the space linearly transformed by $\Omega$: $d_i^L=\left(\Omega\vec{x}_i-\Omega\vec{w}^L\right)^2$, 
\cite{Bunte2011_LiRaMLVQ} used the concept for discriminant visualisation. 
This is achieved by limiting the rank of $\Lambda$ using a rectangular matrix $\Omega\in\R^{M\times D}$ with $M\le D$, which in turn can be used to visualise the piecewise linear decision boundaries if $M\in\{2,3\}$. 


Similarly, to extend ALVQ to global relevances we proposed a parameterized computation of the angle \cite{Bunte_NC_2_2016,ghosh2017comparison}:
\begin{align}
b = b_\Omega = \frac{\vec{x}_i^\top\Omega^\top\Omega\vec{w}^L}
{\lVert\vec{x}_i\rVert_\Omega\lVert\vec{w}^L\rVert}_\Omega \text{ with }
\lVert\vec{v}\rVert_{\Omega}=\sqrt{\vec{v}^\top\Omega^\top\Omega \vec{v}} \enspace ,
\end{align}
with corresponding derivatives:
\begin{align}
\frac{\partial b_\Omega}{\partial \vec{w}^L} &= 
\frac{\vec{x}_i\Omega^\top\Omega\lVert\vec{w}^L\rVert_\Omega^2-\vec{x}_i\Omega^\top\Omega \vec{w}^L\cdot \vec{w}^L\Omega^\top\Omega}{\lVert \vec{x}_i\rVert_\Omega\lVert \vec{w}^L\rVert_\Omega^3}\\ 
\notag\frac{\partial b_\Omega}{\partial \Omega_{md}} =  
&\frac{x_{i,m}\sum_j\Omega_{jd}w^L_j+w^L_m\sum_j\Omega_{jd}x_{i,j}}{\lVert \vec{x}_i\rVert_\Omega\lVert \vec{w}^L\rVert_\Omega} 
- \vec{x}_i\Omega^\top\Omega \vec{w}^L \\
&\cdot\left[
\frac{x_{i,m}\sum_j\Omega_{jd}x_{i,j}}{\lVert\vec{x}_i\rVert^3_\Omega\lVert\vec{w}^L\rVert_\Omega} + 
\frac{w^L_m\sum_j\Omega_{jd}w^L_j}{\lVert\vec{x}_i\rVert_\Omega\lVert \vec{w}^L\rVert^3_\Omega}\right] \enspace, 
\end{align}
where $x_{i,m}$ denotes dimension $m$ of vector $\vec{x}_i$.
As before the diagonal of $\Lambda = \Omega^T\Omega$ denotes the individual feature relevances for the classification and $\Omega$ 
can be rectangular $\Omega \in \R^{M \times D}$ with $M \le D$ to be used for visualisation. 
Resulting visualisations are one $M$ dimensional hyper-spheres where the angle-based classification takes place.
The global Euclidean and angle implementation will be abbreviated by LVQ$_g$ and ALVQ$_g$ respectively. 

\subsection{Local relevance matrix}\label{angleLGMLVQ} 

The localized extension LGMLVQ \cite{schneider2009adaptive} allows more complex modeling and prototype or class-wise feature relevance determination by attaching metric tensors $\Psi^c$ to each prototype or each class 
(based on the user's choice): 
\begin{align} 
d^c_i = (\vec{x}_i-\vec{w}^c)^\top\Psi^{c\top}\Psi^c(\vec{x}_i-\vec{w}^c) \enspace.
\end{align}
This Euclidean variant is powerful for finding solutions to non-linearly separable multi-class problems.
The diagonal of the local metric tensors $\Lambda_c=\Psi^{c\top}\Psi^c$ contain local or class-wise feature relevances, which can be investigated by the user for class-specific discriminative information. 
However, visualising the decision boundaries is not trivial and non-linear mappings based on charting can be found in \cite{Bunte2011_LiRaMLVQ, bunte2010adaptive}.

In this contribution we extend ALVQ learning with missing data to local relevances following similar principles:
\begin{align}
b=b_{\Psi^L} = \frac{\vec{x}_i^\top\Psi^{L\top}\Psi^L\vec{w}^L}
{\lVert\vec{x}_i\rVert_{\Psi^L}\lVert\vec{w}^L\rVert_{\Psi^L}}
\enspace .
\end{align}
The corresponding derivatives of $b_{\Psi^L}$ 
are as follows:
\begin{align}
&\frac{\partial b_{\Psi^L}}{\partial \vec{w}^L} = 
\frac{\vec{x}_i \Psi^{L \top} \Psi^L \lVert \vec{w}^L \rVert_{\Psi^L}^2-\vec{x}_i \Psi^{L\top} \Psi^L \vec{w}^L \cdot \vec{w}^L \Psi^{L\top}\Psi^L}
{\lVert \vec{x}_i\rVert_{\Psi^L}\lVert\vec{w}^L\rVert_{\Psi^L}^3} \\
&\frac{\partial b_{\Psi^L}}{\partial \Psi^L_{md}}  = 
\frac{x_{i,m} \sum_j \Psi_{jd}^L w_{j}^L +w_m^L \sum_j \Psi_{jd}^L x_{i,j}}
{\lVert \vec{x}_i \rVert_{\Psi^L}\lVert \vec{w}^L\rVert_{\Psi^L}} 
- \notag\\
&
\vec{x}_i \Psi^{L\top} \Psi^L w^L 
\left[\frac{x_{i,m} \sum_j \Psi_{jd}^L x_{i,j}}{\lVert \vec{x}_i \rVert^3_{\Psi^L}\lVert \vec{w}^L\rVert_{\Psi^L}} + \frac{w_m^L\sum_j \Psi_{jd}^L w_j^L}
{\lVert \vec{x}_i \rVert_{\Psi^L}\lVert \vec{w}^L\rVert^3_{\Psi^L}}\right]
\end{align}
Similarly to the Euclidean version the local matrices can lead to valuable insight about local or class-wise relevant features and visualisation of the non-linear decision boundaries needs additional effort. 
The local Euclidean and angle implementation will be abbreviated by LVQ$_l$ and ALVQ$_l$ respectively. 

\subsection{2 matrix decomposition for visualisation}\label{angleLLiRAMLVQ} 
As a compromise between linear dimensionality reduction and visualisation of non-linear decision boundaries 
\cite{Bunte2011_LiRaMLVQ} introduced a composition of the matrix in the quadratic form Eq.\ (\ref{eq:d_Lambda}) with two matrices:
\begin{align} 
d^c_i = (\vec{x}_i-\vec{w}^c)^\top\Omega^\top\Psi^{c\top}\Psi^c\Omega(\vec{x}_i-\vec{w}^c) \enspace, 
\end{align}
with $\Omega\in\R^{M\times D}$ and $\Psi^c\in\R^{M \times M}$. 
The data and prototypes are therefore transformed linearly to the $M$-dimensional space and the local metric tensors define the non-linear decision boundaries in that space. 
If the intrinsic dimensionality is more than $M\in\{2,3\}$ a loss of information in classification and visualisation 
is inevitable, however the cost function ensures that this loss is minimized. 

In this contribution we similarly extend 
ALVQ for visualisation with non-linear decision boundaries:
\begin{align}
b=b_{2M} = \frac{\vec{x}_i^\top \Omega^\top \Psi^{L\top} \Psi^L \Omega \vec{w}^{L}}{
\lVert\vec{x}_i\rVert_{2M}\lVert\vec{w}^L\rVert_{2M}
} 
\end{align}
with $\lVert\vec{v}\rv= \sqrt{\vec{v}^\top \Omega^\top \Psi^{L\top} \Psi^L \Omega \vec{v}}$ and derivatives: 
\begin{align}
\notag\frac{\partial b_{2M}}{\partial \vec{w}^L} = 
  &\frac{\vec{x}_i\Omega^\top \Psi^{L\top} \Psi^L \Omega \lVert\vec{w}^L\rv^2}{\lVert\vec{x}_i\rv\lVert\vec{w}^L\rv^3}-  \notag \\
  &\frac{\vec{x}_i\Omega^\top \Psi^{L\top} \Psi^L \Omega \vec{w}^L 
  \cdot \vec{w}^L \Omega^\top \Psi^{L\top} \Psi^L \Omega}{\lVert\vec{x}_i\rv\lVert\vec{w}^L\rv^3} \\
\frac{\partial b_{2M}}{\partial \Omega} =
&\frac{2 \vec{x}_i^\top \Psi^{L\top} \Psi^L \Omega \vec{w}^L}{\lVert \vec{x}_i\rv \lVert \vec{w}^L\rv} 
-\vec{x}_i \Omega^\top \Psi^{L\top} \Psi^L \Omega \vec{w}^L  \cdot \notag\\
& \left[ \frac{\vec{x}_i \Psi^{L\top} \Psi^L \Omega \vec{x}_i}{\lVert \vec{x}_i\rv^3 \lVert \vec{w}^L\rv }  +
\frac{\vec{w}^L \Psi^{L\top} \Psi^L \Omega \vec{w}^L \lVert}{\lVert \vec{x}_i\rv \lVert \vec{w}^L\rv^3 }  \right]  
\\
\frac{\partial b_{2M}}{\partial \Psi^L}  = 
&\frac{2 \vec{x}_i^\top \Omega^\top \Psi^L \Omega \vec{w}^L}{\lVert  \vec{x}_i\rv \lVert \vec{w}^L\rv}
- \vec{x}_i \Omega^\top \Psi^{L\top} \Psi^L \Omega \vec{w}^L \cdot \notag\\
& \left[ \frac{ \vec{x}_i \Omega^\top \Psi^L \Omega \vec{x}_i }{\lVert \vec{x}_i\rv^3 \lVert  \vec{w}^L\rv }  +
\frac{ \vec{w}^L \Omega^\top \Psi^L \Omega \vec{w}^L }{\lVert \vec{x}_i\rv \lVert \vec{w}^L\rv^3 }  \right]
\end{align} 
The 2 matrix Euclidean and angle implementation will be abbreviated by LVQ$_{2M}$ and ALVQ$_{2M}$ respectively.

\section{Datasets}
We demonstrate our newly developed classifiers on two datasets: a synthetic 2-class dataset and a publicly available multi-class heart disease dataset as explained in the following subsections.
\subsection{Synthetic non-linear dataset (Football)}
\begin{figure}[!ht]
\centering
\includegraphics[width=\columnwidth,trim = 1.9cm 1.0cm 1.3cm 1.0cm, clip]{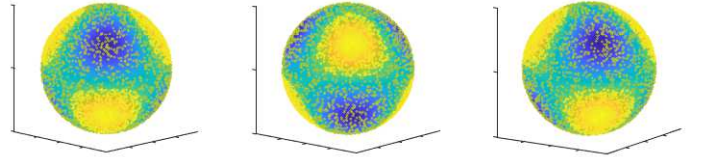}
\caption{Football: 3 different views of a non-linearly separable 
synthetic dataset.}
\label{fig:football}
\end{figure}

We used the open-source software system Chebfun \cite{Chebfun2014} to create a synthetic 2 class dataset resembling the pattern of a football (see Fig.\ref{fig:football}).
The function producing the pattern is $f(\vec{x})=2\sinh(5x_1\cdot x_2\cdot x_3)$ with $f(\vec{x})\le 0.5$ belonging to class 0 and $f(\vec{x})>0.5$ to class 1.
We created 5000 samples for training and validation splits in cross-validation 
and additional 25000 samples serve as hold-out test set to investigate the generalization ability of the classifier. 
The data is available online\footnote{{$ github.com$\slash$sreejita-rug$\slash$Synthetic\_Chebfun\_football.git$}}. 
Performance on this dataset is reported in terms of training and test errors, as well as sensitivity and specificity. 

\subsection{Heart disease dataset from UCI}
This dataset, also known as the Cleveland heart disease (HD) dataset \cite{DavidWAha1988}, contains 303 subjects in total 
(164 healthy, and 139 with varying degrees of heart problems). 
The predictor variable is originally 5 unique values, 0 indicating healthy (164), while 1 (55 subjects), 2 (36 subjects), 3 (35 subjects), and 4 (13 subjects) indicating patients with different heart conditions. 
Furthermore, six subjects contain missing values. 
The dataset originally consists of 76 features but most research has been done on a subset of 13 of these. 
For easy comparison we investigate the same 13 features 
and details about them can be found at the UCI repository \cite{DavidWAha1988}.

%
Exploratory analysis showed that while there is a very good separation 
between healthy and HD subjects considered in binary classification, the multi-class problem differentiating between the 4 classes of HD patients turns out to be remarkably difficult. 
Therefore, besides the more interesting multi-class problem, we added an investigation of the binary sub-problem to compare the performance to the majority of earlier results reported on this dataset. 
However, unlike most contributions we did not discard 
entries with missing values, 
since our method can be trained in variable dimensional spaces. 
According to \cite{DavidWAha1988} the missing values in the data were replaced by a value of -9. For the binary problem we report the performance keeping this, to compare to earlier results. In the multi-class analysis however we revert the -9s to $\mathrm{NaNs}$. 
 

Literature on the heart disease dataset investigating the binary problem, showed good performance by SVMs with non-linear kernels, neural networks, k-nearest neighbour (kNN) using $k=16, 19, 28$,  
Fischer Discriminant Analysis (FDA), Linear Discriminant Analysis (LDA), NB and ensemble classifiers such as RF 
\cite{shouman2012applying, kahramanli2008design, nahar2013computational}. 
Although these classifiers perform well in binary classification of this HD dataset, direct interpretation and visualization of the trained models remains difficult, with exception of the RF. 
Models with enhanced interpretability as proposed in this contribution can alternatively deliver additional insight. This is also demonstrated on the more interesting multi-class problem. 

\section{Experiments}
In this section we explain the experimental setup for the synthetic and heart disease dataset and the performance metrics used for comparison. 
Results are summarized in tables with abbreviations as introduced before:
global feature relevances Euclidean and angle based (LVQ$_g$ and ALVQ$_g$), local relevances (LVQ$_l$ and ALVQ$_l$), Random Forest (RF), and 
the 2 matrix versions providing visualisations of the nonlinear decision boundaries (LVQ$_{2M}$ and ALVQ$_{2M}$)
The superscripts denote the value of hyperparameters $\beta$ for ALVQ and the number of trees in the RF classifier.

\subsection{Synthethic data}
We demonstrate the difference of the localized and 2 matrix Euclidean LVQ versions and our angle based extensions on the synthetic football pattern data set. 
Therefore, we performed a 10-fold cross validation for comparison and model selection with 5000 samples. 
The generalization ability of the selected model is evaluated on 25000 hold-out test samples 
and performance is reported 
in terms of training and test errors, as well as sensitivity and specificity. 
We use 3 prototypes per class ($\wp=6$) and class-wise matrices on this dataset. 

\subsection{Heart disease data}
We compare the proposed angle LVQ variants with results from the literature \cite{nahar2013computational, palaniappan2008intelligent}. 
Contrary to past results our method can perform on the existing dimensions only. This avoids imputation and offers additional insights in the form of feature relevance determination and visualisation of the decision boundaries.
This dataset was z-score transformed in each fold using the mean and standard deviation of the corresponding training set.
Earlier results were typically acquired by 10-fold cross validation, since most of them simplify the problem to two classes, combining all diseases into one. 
However, we use 5-fold cross validation, since the smallest minority class contained only 14 subjects justifying only a lower number of folds for the analysis of the multi-class problem. 
Albeit the multi-class problem being severely more difficult we show that the enhanced interpretability offers additional insight into the problem. 

Table \ref{tab:exp_summary} shows an overview of the experiments performed and intrinsic method hyperparameters. 
The imbalance of the classes is handled by the Synthetic Minority Oversampling TEchnique (SMOTE) as described in \cite{chawla2002smote}. 
SMOTE$^g$ denotes a geodesic variant for oversampling on a hypersphere as introduced and explained in \cite{ghosh2017comparison}. 
They were used to oversample all minority classes in the training set to contain the same number of samples as the majority class (Healthy). Based on exploratory analysis we chose $k=3$ nearest neighbours for both SMOTE and SMOTE$^g$. 
We investigated the intrinsic dimensionality by training full rank matrices $\Omega$ for which subsequent Eigen-value decomposition of the resulting metric tensor $\Lambda$ delivered insight into the 
required dimensions for classification. 
Afterwards we limit the rank for visualisation purpose. 
We experimented with varying number of prototypes per class (1, 2 and 3), such that $\wp\in\{2,4,6\}$ for the binary class problem, and $\wp\in\{5,10,15\}$ for 5-class problem, 
and investigated the influence of the hyperparameter $\beta$ with $\beta\in\{1,5,10,50,80,100\}$.
\begin{table}[t]
\scriptsize
 \caption{Experiments performed on the heart disease dataset.}
\begin{center}
\begin{tabularx}{\columnwidth}{@{\extracolsep{\fill}}|llp{0.19\textwidth}l|} 
\hline
\textbf{classes} & \textbf{Method}	  & \textbf{Hyperparameters}   &   \textbf{Preprocessing} \\
\hline 
Binary 		    & LVQ$_g$	     	  & $\wp$, rank of $\Omega$ & z-score\\
Binary		    & ALVQ$_g$     	  & $\wp$, rank of $\Omega$, $\beta$ & z-score\\
Binary 		    & RF	      	  & No. of trees	 & z-score\\ 
\hline 
5-class	   	    & ALVQ$_g$         	& $\wp$, rank of $\Omega$, $\beta$   & z-score, SMOTE$^g$ \\
5-class	    	    & ALVQ$_l$	 	 & $\wp$, ranks of $\{\Psi_c\}$, $\beta$ 	 & z-score, SMOTE$^g$ \\
5-class	    	    & ALVQ$_{2M}$	  & $\wp$, ranks of $\Omega$\&$\{\Psi_c\}$: , $\wp$, $\beta$    & z-score, SMOTE$^g$\\
5-class	    	    & RF	      	  & No. of trees		 & z-score, SMOTE$^s$ \\
\hline
\end{tabularx}
\label{tab:exp_summary}
\end{center}
\end{table}
As proposed in \cite{breiman2001random} we set minimum observation(s) per tree leaf in RF to $1$, and number of random variables at each decision split to $\sqrt{D} =\sqrt{13}\approx 4 $.

\section{Results and Discussion}
 This section contains the detailed comparison of results from experiments performed on both datasets, followed by discussion and visualizations as enabled by the proposed 
ALVQ$_{2M}$. 
For the real-world heart disease we also showed a detailed investigation of interpretable parameters leading to further insight into the classification performed. 
RFs, which are also interpretable to some extent, makes it possible to extract feature importance. Therefore we were able to compare findings from the ALVQs and RFs.

\subsection{Synthetic football dataset results} 
\begin{figure}[t]
\centering
\includegraphics[width=\columnwidth]{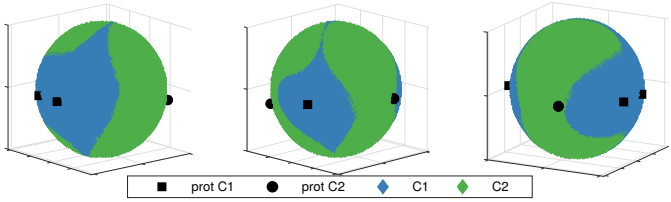}
\caption{Three different perspectives of the nonlinear decision boundaries in the spherical classification space of ALVQ$_{2M}$ trained on the Football dataset.
}
\label{fig:syn_cls}
\end{figure}

Table \ref{tab:syn_cls} summarizes the performance of the classifiers in terms of error on training and test 
set during cross validation and report the sensitivity and specificity with respect to the hold-out test set. 
We included the results using different hyperparameters $\beta$ to provide information about the robustness and selected the model exhibiting best training performance as highlighted in boldface. 
As expected, earlier LVQ extensions perform worse on this non-Euclidean data set as depicted in the first 2 rows. 
The local relevance angle LVQ (ALVQ$_l$) clearly outperforms the other two being the most complex and flexible model with the largest number of parameters handling the nonlinearities of this data best. 
However, as mentioned before, visualization of the decision boundaries with local metric tensors is not straightforward. 
Therefore we demonstrate the 2 matrix extension ALVQ$_{2M}$ with complexity and performance in between the global and local variants. 
\begin{table}[t]
\scriptsize
\caption{Football comparison: mean performance (standard deviation)}
\begin{center}
\begingroup
\begin{tabularx}{\columnwidth}{@{\extracolsep{\fill}}|lrrrr|} 
\hline
\textbf{Method} &\multicolumn{1}{l}{\textbf{$E_\mathrm{train}$}}& \multicolumn{1}{l}{\textbf{$E_\mathrm{test}$}}& \multicolumn{1}{l}{\textbf{Sensitivity}}& \multicolumn{1}{l|}{\textbf{Specificity}} \\
\hline
LVQ$_{2M}$ & 0.272 (0.019) 	& 0.277 (0.027) & 0.68 (0.093)		& 0.76 (0.103) \\
LVQ$_l$ & 0.223 (0.047) 	& 0.224 (0.050) & 0.78 (0.118)		& 0.76 (0.113) \\
\hline  
{$\mathbf{ALVQ_{\scaleto{g}{3pt}}^{\scaleto{10}{2.5pt}}}$}  & \textbf{0.268 (0.035)}& 0.273 (0.036)& \textbf{0.78 (0.115)}	& 0.67 (0.103) \\
$ALVQ_{\scaleto{g}{3pt}}^{\scaleto{30}{2.5pt}}$  & 0.273 (0.040) 	& 0.285 (0.047) & 0.76 (0.127)		& 0.68 (0.115) \\
$ALVQ_{\scaleto{g}{3pt}}^{\scaleto{50}{2.5pt}}$  & 0.271 (0.040) 	& 0.279 (0.041) & 0.76 (0.118)		& 0.69 (0.111) \\
$ALVQ_{\scaleto{g}{3pt}}^{\scaleto{80}{2.5pt}}$  & 0.284 (0.048) 	& 0.290 (0.051) & 0.74 (0.139)		& 0.68 (0.145) \\
$ALVQ_{\scaleto{g}{3pt}}^{\scaleto{100}{2.5pt}}$ & 0.277 (0.042) 	& 0.288 (0.046) & 0.75 (0.130)		& 0.68 (0.135) \\
$ALVQ_{\scaleto{g}{3pt}}^{\scaleto{120}{2.5pt}}$ & 0.286 (0.047) 	& 0.298 (0.048) & 0.74 (0.123)		& 0.66 (0.123) \\
\hline
$ALVQ_{\scaleto{l}{3pt}}^{\scaleto{10}{2.5pt}}$ & 0.199 (0.056) 	& 0.202 (0.060) & 0.82 (0.117)		& 0.77 (0.111) \\
{$\mathbf{ALVQ_{\scaleto{l}{3pt}}^{\scaleto{30}{2.5pt}}}$} & \textbf{0.176 (0.066)}& 0.182 (0.070) & \textbf{0.82 (0.140)} & 0.82 (0.113) \\
$ALVQ_{\scaleto{l}{3pt}}^{\scaleto{50}{2.5pt}}$ & 0.197 (0.059) 	& 0.204 (0.064) & 0.79 (0.144)		& 0.80 (0.117) \\
$ALVQ_{\scaleto{l}{3pt}}^{\scaleto{80}{2.5pt}}$ & 0.196 (0.062) 	& 0.208 (0.064) & 0.79 (0.129)		& 0.80 (0.108) \\
$ALVQ_{\scaleto{l}{3pt}}^{\scaleto{100}{2.5pt}}$& 0.191 (0.059) 	& 0.200 (0.060) & 0.79 (0.141)		& 0.82 (0.100) \\
$ALVQ_{\scaleto{l}{3pt}}^{\scaleto{120}{2.5pt}}$& 0.201 (0.057) 	& 0.208 (0.061) & 0.77 (0.142)		& 0.81 (0.110) \\
\hline
$ALVQ_{\scaleto{2M}{3pt}}^{\scaleto{10}{2.5pt}}$ & 0.24 (0.057)         & 0.24 (0.059) & 0.80 (0.124)       & 0.72 (0.116) \\ 
$ALVQ_{\scaleto{2M}{3pt}}^{\scaleto{30}{2.5pt}}$ & 0.23 (0.052)         & 0.23 (0.056) & 0.77 (0.140)        & 0.76 (0.122) \\
{$\mathbf{ALVQ_{\scaleto{2M}{3pt}}^{\scaleto{50}{2.5pt}}}$} & \textbf{0.22 (0.058)}& 0.22 (0.061) & \textbf{0.79 (0.133)} & 0.75 (0.117) \\
$ALVQ_{\scaleto{2M}{3pt}}^{\scaleto{80}{2.5pt}}$ & 0.24 (0.058)         & 0.25 (0.062) & 0.76 (0.146)       & 0.74 (0.136)  \\
$ALVQ_{\scaleto{2M}{3pt}}^{\scaleto{100}{2.5pt}}$& 0.24 (0.058)         & 0.24 (0.060) & 0.78 (0.140)       & 0.73 (0.130)  \\
$ALVQ_{\scaleto{2M}{3pt}}^{\scaleto{120}{2.5pt}}$& 0.24 (0.061)         & 0.24 (0.063) & 0.77 (0.141)       & 0.73 (0.140)\\
\hline
\end{tabularx}
\endgroup
\label{tab:syn_cls}
\end{center}
\end{table}
Figure \ref{fig:syn_cls} shows a corresponding example visualization of the nonlinear decision boundaries and prototypes in the spherical classification space seen from 3 different perspectives. 
Individual data samples have been omitted in the illustration to reduce visual clutter, but can be added of course for investigation. 
%

\subsection{Heart disease (binary class problem) results}
\begin{figure}[!t]
\centering
\includegraphics[width=\columnwidth]{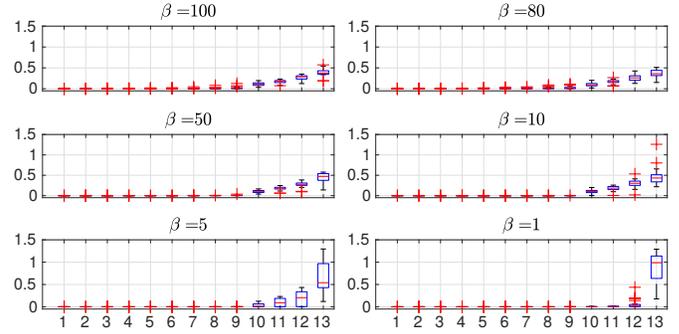}
 \caption{Eigenvalues of $\Lambda$ across 5 folds and 5 runs.}
 \label{fig:evd_hd}
\end{figure}

First, we investigated the binary subproblem combining all diseases to one class and estimate the intrinsic dimensionality by investigating the eigenvalue profile 
of the trained $\Lambda=\Omega^\top\Omega$ with full rank $\Omega\in\R^{D\times D}$ and one prototype per class. 
Since there is not enough data to create a hold-out generalization set we report 
the sensitivity and specificity of the classifiers as observed on the test set of the cross-validation splits. 
Figure \ref{fig:evd_hd} shows box plots of the estimates of the intrinsic dimensionality according to different settings of the hyperparameter $\beta$ 
and the average performance of corresponding models is summarized in Table \ref{tab:hd_2cls13dim}.
\begin{table}[t!]
\scriptsize
\caption{binary HD: mean performance (std) of global full rank ALVQ}
\begin{center}
\begingroup
\renewcommand{\arraystretch}{1.1}
\begin{tabularx}{\columnwidth}{@{\extracolsep{\fill}}|lrrrr|} 
\hline
\textbf{Method} &\multicolumn{1}{l}{\textbf{$E_\mathrm{train}$}}& \multicolumn{1}{l}{\textbf{$E_\mathrm{test}$}}& \multicolumn{1}{l}{\textbf{Sensitivity}}& \multicolumn{1}{l|}{\textbf{Specificity}} \\
\hline
$ALVQ_{\scaleto{g}{3pt}}^{\scaleto{1}{2.5pt}}$  & 0.112 (0.009) 	& 0.171 (0.029) & 0.79 (0.084)		& 0.86 (0.094) \\
{$\mathbf{ALVQ_{\scaleto{g}{3pt}}^{\scaleto{5}{2.5pt}}}$}& \textbf{0.110 (0.010)}  & 0.178 (0.044) & \textbf{0.78 (0.097)} & 0.86 (0.086) \\
$ALVQ_{\scaleto{g}{3pt}}^{\scaleto{10}{2.5pt}}$& 0.112 (0.015)		& 0.188 (0.040) & 0.78 (0.120)		& 0.84 (0.098) \\
$ALVQ_{\scaleto{g}{3pt}}^{\scaleto{50}{2.5pt}}$& 0.130 (0.018) 		& 0.181 (0.044) & 0.80 (0.066)		& 0.83 (0.089) \\
$ALVQ_{\scaleto{g}{3pt}}^{\scaleto{80}{2.5pt}}$& 0.133 (0.019) 		& 0.180 (0.046) & 0.80 (0.076)		& 0.84 (0.090) \\
$ALVQ_{\scaleto{g}{3pt}}^{\scaleto{100}{2.5pt}}$& 0.140 (0.025) 	& 0.202 (0.050) & 0.77 (0.088)		& 0.82 (0.081) \\
\hline
\end{tabularx}
\endgroup
\label{tab:hd_2cls13dim} 
\end{center}
\end{table}
Even though there are 13 features in the dataset much lower dimensionality seems necessary for classification as indicated by most Eigenvalues being close to 0. 
The best performing $\beta$ depicts only three Eigenvalues significantly bigger than 0 indicating the problem can be visualized in three dimensions with limited loss of information. Thus we re-train the models by limiting the rank to three ($M=3$). 
\begin{table}[ht!]
\scriptsize
\caption{
binary HD: mean performance (std) final comparison}
\begin{center}
\begingroup
\renewcommand{\arraystretch}{1.2}
\begin{tabularx}{\columnwidth}{@{\extracolsep{\fill}}|lllll|}
\hline 
\textbf{Method} &\multicolumn{1}{l}{\textbf{$E_{train}$}}& \multicolumn{1}{l}{\textbf{$E_{test}$}}& \multicolumn{1}{l}{\textbf{Sensitivity}}& \multicolumn{1}{l|}{\textbf{Specificity}} \\
\hline
$LVQ_g$    & 0.459 (0.001) & 0.459 (0.005) & 0.00 (0.000)& 1.00 (0.000) \\
\hline
{$\mathbf{ALVQ_{\scaleto{g}{3pt}}^{\scaleto{1}{3pt}}}$}  & \textbf{0.114 (0.009)} & 0.169 (0.034) & \textbf{0.81 (0.077)}& 0.85 (0.098) \\
$ALVQ_{\scaleto{g}{3pt}}^{\scaleto{5}{2.5pt}}$  & 0.116 (0.012) & 0.178 (0.048) & 0.79 (0.110)& 0.85 (0.088) \\
$ALVQ_{\scaleto{g}{3pt}}^{\scaleto{10}{2.5pt}}$ & 0.122 (0.013) & 0.187 (0.050) & 0.79 (0.101)& 0.83 (0.095) \\
$ALVQ_{\scaleto{g}{3pt}}^{\scaleto{50}{2.5pt}}$& 0.145 (0.023) & 0.199 (0.044) & 0.80 (0.081)& 0.80 (0.074) \\
$ALVQ_{\scaleto{g}{3pt}}^{\scaleto{80}{2.5pt}}$ & 0.168 (0.029) & 0.210 (0.052) & 0.76 (0.096)& 0.81 (0.080) \\
$ALVQ_{\scaleto{g}{3pt}}^{\scaleto{100}{2.5pt}}$& 0.163 (0.033) & 0.204 (0.052) & 0.79 (0.075)& 0.80 (0.066) \\
\hline
$RF^{\scaleto{50}{3pt}}$    & 0.001 (0.002) & 0.179 (0.044) & 0.78 (0.084)          & 0.86 (0.058) \\
{$\mathbf{RF^{\scaleto{100}{3pt}}}$}   & \textbf{0.0 (0.0)} & 0.179 (0.044) & \textbf{0.77 (0.082)}& 0.86 (0.048) \\
{$\mathbf{RF^{\scaleto{150}{3pt}}}$}   & \textbf{0.0 (0.0)} & 0.170 (0.043) & \textbf{0.78 (0.081)}& 0.87 (0.046) \\
{$\mathbf{RF^{\scaleto{200}{3pt}}}$}   & \textbf{0.0 (0.0) }& 0.177 (0.050) & \textbf{0.77 (0.082)}& 0.86 (0.053) \\
 \hline
$NB^{\scaleto{Kol}{3pt}}$           & NA       & NA          &0.86               & 0.833 \\
$MLP^{\scaleto{Kol}{3pt}}$          & NA       & NA          & 0.836             & 0.80\\
\hline
\end{tabularx}
\endgroup
\label{tab:hd_2cls3dim}
\end{center}
\end{table}

As before we perform model selection and highlight in boldface based on the best training set performance and report sensitivity and specificity on the respective test splits. 
Reducing the rank of the matrix regularizes the model leading to improved generalization performance as depicted in Table \ref{tab:hd_2cls3dim}. 
We also notice that the Euclidean versions of LVQ, i.e. GMLVQ, exhibits poor performance on this data set. 
This might be due to the presence of missing data, which the angle version is able to deal with. 
RF with 100 and more trees have had perfect training, but the sensitivity on the validation set is similar to that of angle LVQ. 
We also observe similar performance in comparison with results reported in \cite{kahramanli2008design} for 
the NB and Multi Layer Perceptron (MLP) marked as NB$^{Kol}$ and MLP$^{Kol}$. 
They used 10-fold cross-validation, but standard deviation across the different splits or training and test error were not reported. 

\begin{figure}
\centering
\includegraphics[width=\columnwidth]{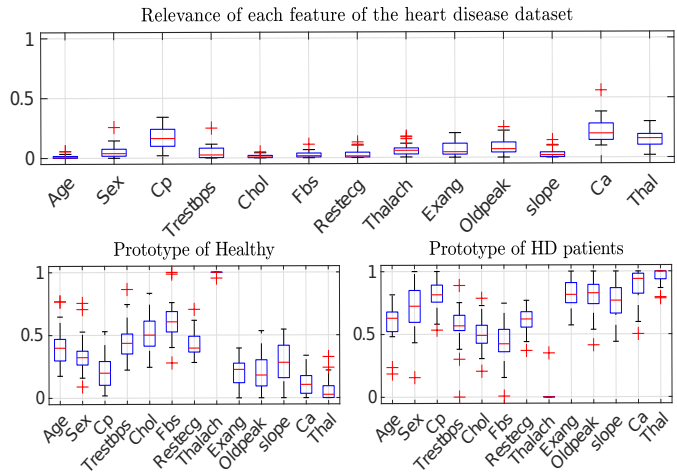}
\caption{Feature relevances (top panel), as well as healthy and disease prototypes (bottom row) obtained by ALVQ$_g^1$ on the binary HD classification.
}
\label{fig:hd_AGMLVQ}
\end{figure}

Classifiers of the LVQ family can also identify relevant features for a particular task, along with finding typical representatives of each class (prototypes).
Figure \ref{fig:hd_AGMLVQ} shows the feature relevances and prototypes of the healthy and disease class learned during training corresponding to the best setting  ($ALVQ_g^1$) in Table \ref{tab:hd_2cls3dim}. 
%
\begin{figure}
\includegraphics[width=\columnwidth]{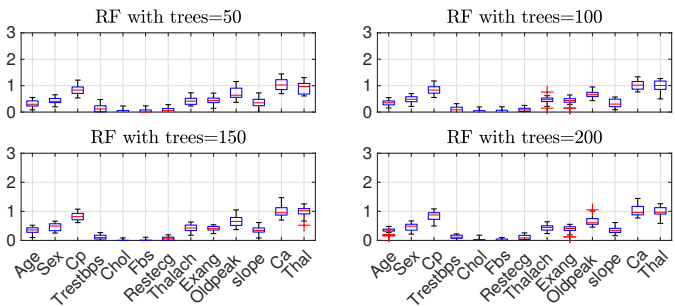}
\caption{Summary of the feature importance determined by RF over 5 folds and 5 runs, trained for the binary class problem.}
\label{fig:hd_RF}
\end{figure}
The features 3 (Chest pain type), 12 (number of major vessels as coloured by fluoroscopy) and 13 
(status of heart, w.r.t the organ being normal having had the anomaly fixed, and having a reversible defect) are among the most highly relevant ones, 
followed by features 2 (sex), 8 (maximum heart rate achieved) and 9 (exercise induced angina). 
Important features extracted from RF models are shown in Fig. \ref{fig:hd_RF}. 
RF and ALVQ feature sets agree with regard to features 3, 8, 9, 10, 12, and 13
being the more distinguishing ones, whereas features 4, 5 and 7 do not contribute as much. 
In contrast to RF we can also investigate the prototypes of the healthy and patient class. 
Notably, the features found also visibly differ in the adapted prototypes of the healthy and patient. 
We see in Figure \ref{fig:hd_AGMLVQ}, that feature 3 value lie below the 0.5 mark for class-Healthy whereas it is higher that than 0.5 mark for the HD prototype. 
Similarly for value of features 12, 13, 2, and 9. 
For features 8 and 10 the opposite trend is seen: the maximum heart rate achieved for the prototype describing the healthy subjects was much higher than the 0.5 mark 
whereas for the patients it was significantly lower than that mark. 
Conversely, features which were not deemed highly relevant by our classifier, such as features 1 (age), 4 (resting blood pressure), and 7 (resting ECG), 
are seen to have values in the mid-part of the prototype plots for both the classes, 
thus indicating that they are not as integral to distinguishing between healthy subjects and HD patients. 
These findings also agree with those mentioned in \cite{nahar2013computational, palaniappan2008intelligent}. 
%

\subsection{Heart disease (5-class problem) results} 
\begin{figure}
\centering
\includegraphics[width=\columnwidth]{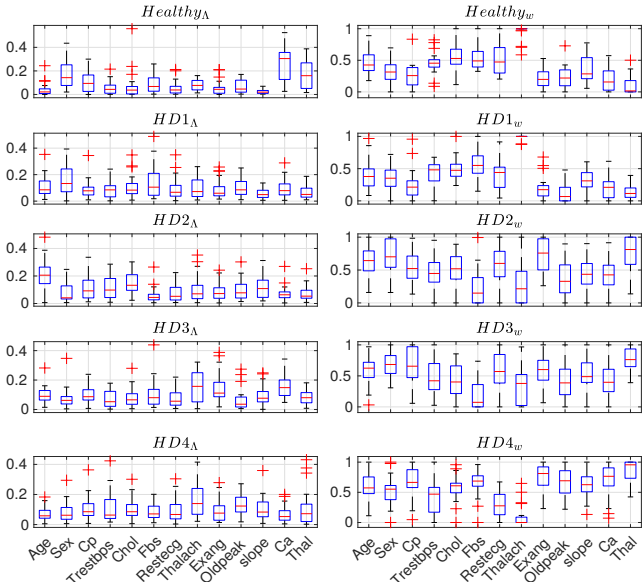}
\caption{Local relevances (left) and prototypes (right) of Healthy, and HD-patients of types 1-4, from ALVQ$_l$ over 5 folds, for the 5-class problem. 
}
\label{fig:hd_ALGMLVQ}
\end{figure}

More challenging and potentially more interesting is the investigation of the 5 class problem keeping the original HD sub-classes. 
Since there are 5 classes we show the performance in terms of training and test errors, and class-wise accuracies. 
The class-wise accuracy of the Healthy class (C0) is the same as specificity and therefore omitted in the following. 
\begin{table*}
\scriptsize
\caption{
5-class HD: mean performance (std) comparison of ALVQ variants and RF
}
\begin{center}
\begingroup
\renewcommand{\arraystretch}{1.1}
\begin{tabular}{
|llrrrllll|
}
\hline
\textbf{Method} &\textbf{$E_{train}$}     &\multicolumn{1}{l}{\textbf{$E_{test}$}}& \multicolumn{1}{l}{\textbf{Sens}}& \multicolumn{1}{l}{\textbf{Spec}}& \multicolumn{1}{l}{\textbf{C1}}& \multicolumn{1}{l}{\textbf{C2}} & \multicolumn{1}{l}{\textbf{C3}}& \multicolumn{1}{l|}{\textbf{C4}}\\
\hline
$ALVQ_{\scaleto{g}{3pt}}^{\scaleto{100}{2.5pt}}$     & 0.34 (0.032) 		& 0.54 (0.081) & 0.07 (0.043)& 0.68 (0.130) & 0.19 (0.111) 	   & 0.22 (0.209)	  & 0.20 (0.180) 	  & 0.25 (0.260)\\
$ALVQ_{\scaleto{g}{3pt}}^{\scaleto{80}{2.5pt}}$    & 0.33 (0.034) 		& 0.52 (0.075) & 0.08 (0.048)& 0.69 (0.090) & 0.21 (0.125) 	   & 0.23 (0.167)	  & 0.26 (0.188) 	  & 0.29 (0.298)\\
$ALVQ_{\scaleto{g}{3pt}}^{\scaleto{150}{2.5pt}}$     & \textbf{0.32 (0.043)} 	& 0.53 (0.061) & 0.08 (0.053)& 0.71 (0.090) & \textbf{0.20 (0.134)}& \textbf{0.18 (0.148)}& \textbf{0.22 (0.143)} &\textbf{0.13 (0.204)}\\
$ALVQ_{\scaleto{g}{3pt}}^{\scaleto{10}{2.5pt}}$     & 0.35 (0.071) 		& 0.48 (0.056) & 0.08 (0.060)& 0.76 (0.063) & 0.21 (0.154) 	   & 0.21 (0.168)	  & 0.26 (0.196) 	  & 0.23 (0.281)\\
$ALVQ_{\scaleto{g}{3pt}}^{\scaleto{5}{2.5pt}}$       & 0.35 (0.071) 		& 0.50 (0.074) & 0.04 (0.045)& 0.77 (0.097) & 0.11 (0.112) 	   & 0.19 (0.153)	  & 0.26 (0.210) 	  & 0.29 (0.313)\\
$ALVQ_{\scaleto{g}{3pt}}^{\scaleto{1}{2.5pt}}$        & 0.37 (0.063) 		& 0.49 (0.053) & 0.05 (0.049)& 0.79 (0.088) & 0.12 (0.125) 	   & 0.19 (0.167)	  & 0.25 (0.161) 	  & 0.22 (0.288)\\
\hline
$ALVQ_{\scaleto{l}{3pt}}^{\scaleto{100}{2.5pt}}$    & 0.24 (0.048) 		& 0.51 (0.067) & 0.08 (0.060)& 0.69 (0.090) & 0.20 (0.155) 	   & 0.31 (0.131)	  & 0.34 (0.193) 	  & 0.13 (0.226)\\
$ALVQ_{\scaleto{l}{3pt}}^{\scaleto{80}{2.5pt}}$     & 0.21 (0.034) 		& 0.49 (0.049) & 0.09 (0.071)& 0.73 (0.066) & 0.22 (0.186) 	   & 0.31 (0.134)	  & 0.28 (0.208)  	  & 0.15 (0.240)\\
$ALVQ_{\scaleto{l}{3pt}}^{\scaleto{50}{2.5pt}}$     & 0.18 (0.038) 		& 0.49 (0.062) & 0.09 (0.066)& 0.72 (0.061) & 0.22 (0.168) 	   & 0.26 (0.152)	  & 0.33 (0.164) 	  & 0.17 (0.276)\\
$ALVQ_{\scaleto{l}{3pt}}^{\scaleto{10}{2.5pt}}$     & \textbf{0.16 (0.025)} 	& 0.48 (0.049) & 0.08 (0.065)& 0.76 (0.065) & \textbf{0.20 (0.168)}& \textbf{0.28 (0.173)}& \textbf{0.31 (0.149)} & \textbf{0.05 (0.150)}\\
$ALVQ_{\scaleto{l}{3pt}}^{\scaleto{5}{2.5pt}}$       & \textbf{0.16 (0.025)} 	& 0.49 (0.052) & 0.07 (0.061)& 0.74 (0.084) & \textbf{0.17 (0.159)}& \textbf{0.30 (0.206)}& \textbf{0.34 (0.179)} & \textbf{0.05 (0.132)}\\
$ALVQ_{\scaleto{l}{3pt}}^{\scaleto{1}{2.5pt}}$        & 0.17 (0.022) 		& 0.50 (0.056) & 0.07 (0.053)& 0.74 (0.089) & 0.18 (0.138) 	   & 0.23 (0.155)	  & 0.31 (0.189) 	  & 0.08 (0.167)\\
\hline
$ALVQ_{\scaleto{2M}{3pt}}^{\scaleto{100}{2.5pt}}$   & 0.38 (0.064) 		& 0.53 (0.071) & 0.09 (0.066)& 0.68 (0.109) & 0.23 (0.172) 	   & 0.22 (0.190)	  & 0.22 (0.160) 	  & 0.27 (0.315)\\
$ALVQ_{\scaleto{2M}{3pt}}^{\scaleto{80}{2.5pt}}$ & 0.36 (0.058)		& 0.55 (0.074) & 0.06 (0.053)& 0.67 (0.121) & 0.15 (0.136) 	   & 0.21 (0.149)	  & 0.25 (0.212) 	  & 0.23 (0.308)\\
$ALVQ_{\scaleto{2M}{3pt}}^{\scaleto{50}{2.5pt}}$   & 0.35 (0.059) 		& 0.51 (0.075) & 0.07 (0.063)& 0.70 (0.122) & 0.18 (0.161) 	   & 0.21 (0.154)	  & 0.35 (0.207) 	  & 0.21 (0.232)\\
$ALVQ_{\scaleto{2M}{3pt}}^{\scaleto{1}{2.5pt}}$ & 0.34 (0.050) 		& 0.51 (0.063) & 0.07 (0.046)& 0.72 (0.098) & 0.17 (0.117) 	   & 0.24 (0.177)	  & 0.26 (0.160) 	  & 0.24 (0.268)\\
$ALVQ_{\scaleto{2M}{3pt}}^{\scaleto{5}{2.5pt}}$     & \textbf{0.31 (0.033)} 	& 0.49 (0.073) & 0.06 (0.052)& 0.76 (0.108) & \textbf{0.16 (0.133)}& \textbf{0.26 (0.168)}& \textbf{0.31 (0.198)} & \textbf{0.17 (0.252)}\\
$ALVQ_{\scaleto{2M}{3pt}}^{\scaleto{1}{2.5pt}}$ & 0.32 (0.043) 		& 0.49 (0.072) & 0.06 (0.050)& 0.75 (0.089) & 0.15 (0.129) 	   & 0.26 (0.180)	  & 0.30 (0.216) 	  & 0.30 (0.337)\\
\hline
$RF^{50}$          & 0.0 (0.002)  		& 0.46 (0.039) & 0.06 (0.044)& 0.85 (0.054) & 0.15 (0.112) 	   & 0.31 (0.110)	  & 0.15 (0.095) 	  & 0.06 (0.134)\\
$RF^{100}$         & \textbf{0.0 (0.0)}  		& 0.46 (0.030) & 0.03 (0.028)& 0.88 (0.060) & \textbf{0.07 (0.069)}& \textbf{0.30 (0.164)}& \textbf{0.09 (0.071)} &\textbf{0.0 (0.0)}\\
$RF^{150 }$        & \textbf{0.0 (0.0)}		& 0.44 (0.022) & 0.05 (0.036)& 0.88 (0.049) & \textbf{0.13 (0.095)}& \textbf{0.28 (0.120)}& \textbf{0.23 (0.117)} & \textbf{0.0 (0.0)}\\
$RF^{200}$         & \textbf{0.0 (0.0)}		& 0.45 (0.020) & 0.04 (0.039)& 0.87 (0.049) & \textbf{0.09 (0.102)}& \textbf{0.33 (0.144)}&\textbf{0.20 (0.117)} & \textbf{0.10 (0.204)}\\
\hline
\end{tabular}
\endgroup
\label{tab:hd_5cls3dim}
\end{center}
\end{table*}
Table \ref{tab:hd_5cls3dim} shows that the class-wise accuracies during validation are better from the more complex local model of angle LVQ (ALVQ$_l$),
whose prototypes and local relevances are depicted in Figure \ref{fig:hd_ALGMLVQ}. 
Additional interpretation can be gained by using the proposed 2 matrix variant ALVQ$_{2M}$. 
For this problem we compared using 1, 2 and 3 prototypes per class but report only the results using 2 prototypes per class, since it depicted the 
best averaged class-wise accuracy on training. 
The study in \cite{nahar2013computational} attempts to investigate the disease classes 
considering one class versus all classification. 
Their highest sensitivity per condition in this simplified setting were reported to be: 0.891 (Healthy, Sequential minimal optimization (SMO)), 0.321 (HD class 1, IBK from Weka), 0.405 (HD class 2, NB), 0.472 (HD class 3, NB) and 0.214 (HD class 4, IBK) furthermore confirming the complexity of the multi-class problem we investigate. 
Table \ref{tab:hd_5cls3dim} shows that the performance of RF and the ALVQ classifiers were comparable in sensitivity and specificity in the more complex 5-class setting. 
However the ALVQ models can provide additional insight by prototypes and visualizations.

\begin{figure}
\centering
\includegraphics[width=0.9\columnwidth, trim = 0.5cm 0.5cm 0.0cm 0.3cm,clip]{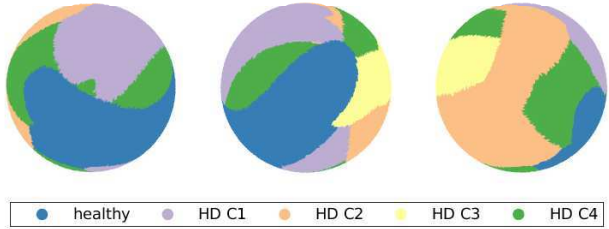}
\caption{Three example perspectives of the classification sphere depicting the decision boundaries as determined by ALVQ$_{2M}$ on the 5-class HD problem.
}
\label{fig:hd_5cls}
\end{figure}
\begin{figure}
  \centering
\includegraphics[width=\columnwidth]{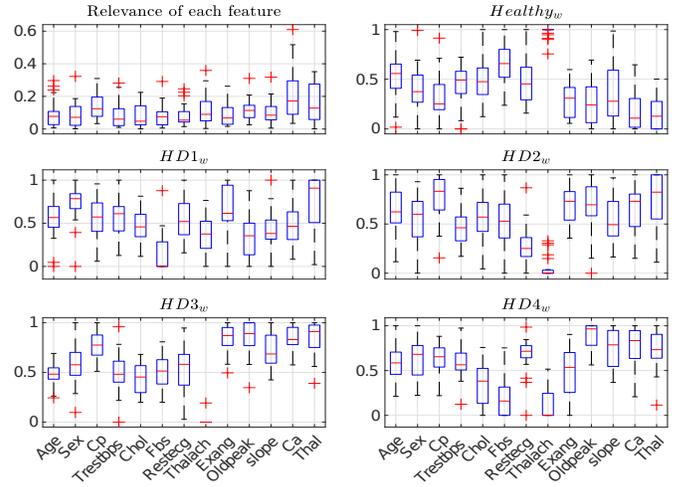}
\caption{
Global relevance, and prototypes ($w$) of Healthy and the 4 types of HD patients, from $ALVQ^5_{2M}$, over 5 folds, trained for the 5-class problem.
}
\label{fig:hd_ALiRaMLVQ}
\end{figure}

Figure \ref{fig:hd_5cls} shows the decision boundaries of an example ALVQ$_{2M}$ using $\beta=5$ showing best performance according to Table \ref{tab:hd_5cls3dim}. 
The picture confirms the non-linearity of this dataset when investigated as 5-class problem. 
Individual data samples are again omitted to avoid visual clutter but can be added and investigated with respect to their distance to the decision boundaries. 
The corresponding cross-validation relevances of $\Omega^\top\Omega$ and prototypes are depicted in Figure \ref{fig:hd_ALiRaMLVQ}. 
Next we investigate the models trained for the multi-class problem in more detail to hypothesize why this problem is so difficult.
Figure \ref{fig:hd_ALGMLVQ} illustrates $ALVQ_l$ classifier with $\beta=5$ and $\Psi_c$ of dimension $3 \times 13$, the hyperparameter setting which showed best performance among angle local LVQ according to Table \ref{tab:hd_5cls3dim}. 
We compare the feature relevance from $\Psi_c$ (Fig. \ref{fig:hd_ALGMLVQ}) with those from figures \ref{fig:hd_AGMLVQ} and \ref{fig:hd_RF}. 
Features 3, 12, and 13 were among the most relevant features for the binary class problem. 
However, for the multi-class problem, on checking the prototype of each class, 
we see that these features do not have a distinct value boundary which could help in identification of the different classes. 
If we consider feature 12 (Ca) for all the prototypes we can see how easily healthy subjects and patients from class sick-1 would be confused, 
similarly patients of sick class 2 would be easily confused with those from sick class 3. 
According to Fig.\ \ref{fig:hd_ALiRaMLVQ} features 12 (Ca) and 13 (Thal) are still the most relevant ones. 
However the prototypes show that these features are good for distinguishing between healthy and the rest of the classes, but not that efficient for differentiating 
between the heart disease classes themselves. 
These plots further explain why the specificity (or the class-wise accuracy of Healthy class) remained high even for the multi-class problem, 
whereas the class-wise accuracies were comparably poor for the remaining. 

\section{Conclusion and future Work}
In this contribution we proposed three interpretable extensions of the angular nearest prototype based classifier, 
namely global angle LVQ, local angle LVQ and a 2 matrix version allowing visualisation of the non-linear decision boundaries. 
These set of classifiers are able to handle missingness as well as make knowledge extraction straightforward. 
As increasing number of human-centric sectors are becoming dependent on machine learning, understanding the exact working and underlying mechanisms behind a decision made by a model, 
are becoming paramount. 
Some classifiers depict comparable (and some even slightly higher) performance than these newly introduced classifiers. 
However, the proposed classifiers captivate due to their interpretability and the 
possibility to shed light on what exactly makes a classification problem difficult. 
This is highlighted in the given analysis of the 5 class heart disease identification problem where we achieve comparable performance to the RF.
Even though the 13 out of 76 features were capable of distinguishing between healthy and heart disease patients, 
but features which can differentiate between all these 5 classes satisfactory seem not to be among these 14 features. 
Future contributions should investigate the larger feature set and the insight we can gain from it using interpretable classifiers. 

\section*{Acknowledgment}
 \footnotesize{
We thank the Center for Information Technology of the University of Groningen for their support and for providing access to the Peregrine high performance computing cluster. We also thank the Rosalind Franklin fellowship, co-funded by the European Union's Seventh Framework Program for research, technological development and demonstration under grant agreement no \textbf{600211}, and H2020-MSCA-IF-2014, project ID \textbf{659104}.
 }

\bibliographystyle{unsrt}
\bibliography{arxiv}

\end{document}